\pdfoutput=1

\documentclass[11pt]{article}
\usepackage[final]{24acl/acl}

\usepackage{times}
\usepackage{latexsym}
\usepackage[T1]{fontenc}

\usepackage[utf8]{inputenc}

\usepackage{microtype}

\usepackage{inconsolata}

\usepackage{dsfont}
\usepackage{xcolor}
\usepackage{subcaption}
\usepackage{float}
\usepackage{multirow}
\usepackage{array}
\usepackage{verbatim}
\usepackage{soul}
\usepackage{url}
\usepackage{graphicx}
\usepackage{amsmath}
\usepackage{amsthm}
\usepackage{amsfonts}
\usepackage{booktabs}
\usepackage{algorithm}
\usepackage{algorithmic}
\usepackage[switch]{lineno}
\usepackage{tcolorbox}

\usepackage{booktabs} 
\usepackage[table]{xcolor} 
\usepackage{hyperref}

\title{The Overlooked Repetitive Lengthening Form in Sentiment Analysis}

\author{Lei Wang \and Eduard Dragut\\
  Temple University, Philadelphia, PA, USA \\
  \texttt{\{tom.lei.wang,edragut\}@temple.edu} \\
 }

\begin{document}
\maketitle
\begin{abstract}
Individuals engaging in online communication frequently express personal opinions with informal styles (e.g., memes and emojis).
While Language Models (LMs) with informal communications have been widely discussed, a unique and emphatic style, the Repetitive Lengthening Form (RLF), has been overlooked for years.
In this paper, we explore answers to two research questions: 1) Is RLF important for sentiment analysis (SA)? 2) Can LMs understand RLF?
Inspired by previous linguistic research, we curate \textbf{Lengthening}, the first multi-domain dataset with 850k samples focused on RLF for SA.
Moreover, we introduce \textbf{Exp}lainable \textbf{Instruct}ion Tuning (\textbf{ExpInstruct}), a two-stage instruction tuning framework aimed to improve both performance and explainability of LLMs for RLF. 
We further propose a novel unified approach to quantify LMs'  understanding of informal expressions.
We show that RLF sentences are expressive expressions and can serve as signatures of document-level sentiment. Additionally, RLF has potential value for online content analysis.
Our results show that fine-tuned Pre-trained Language Models (PLMs) can surpass zero-shot GPT-4 in performance but not in explanation for RLF.
Finally, we show ExpInstruct can improve the open-sourced LLMs to match zero-shot GPT-4 in performance and explainability for RLF with limited samples. Code and sample data are available at \url{https://github.com/Tom-Owl/OverlookedRLF}
\end{abstract}

\section{Introduction}
Informal styles are prevalent on social media platforms, where people use nuanced expressions to share opinions and emotions personally and engagingly \cite{yang-2020-predicting-personal,hosseinia-etal-2021-usefulness, Lihong_dynamics_2019,He_Shen_Mukherjee_Vucetic_Dragut_2021}.
Previous research has explored various informal styles such as meme \cite{lin2024goat,Sharma2023WhatDYMeMe}, 
emoji \cite{peng2023emojilm,barbieri2018multimodal,reelfs2022InterpretingEmoji}, 
slogan \cite{Iwama2018JapaneseSlogan,Misawa2020DistinctiveSlogan} 
and abbreviation \cite{Gorman2021StructuredAbbreviation}.
However, it remains a challenge for Language Models (LMs) to understand the sentiment in nuanced and subtle linguistic expressions, which require a deep contextual and cultural understanding \cite{zhang2023sentiment}.
This work delves into one specific informal expression - Repeated Lengthening Form (RLF), which refers to the linguistic phenomenon where additional characters are added to the standard spelling of a word to enhance or alter its conveyed meaning \cite{Brody2011cool,KALMAN2014187}. 
We further generalize the concept of RLF and divide it into two types: Repetitive Letters (e.g., 'loooove') and Repetitive Punctuations (e.g., 'love!!!!'). 
Our study finds that an average of 5.8\% documents possess RLF among 4 public datasets and 5 domains (Table \ref{tab:dataset_overview}) where some of them include more than 13\% documents with RLF. 

\begin{figure*}[ht]
\centering
\includegraphics[width=1.0\textwidth]{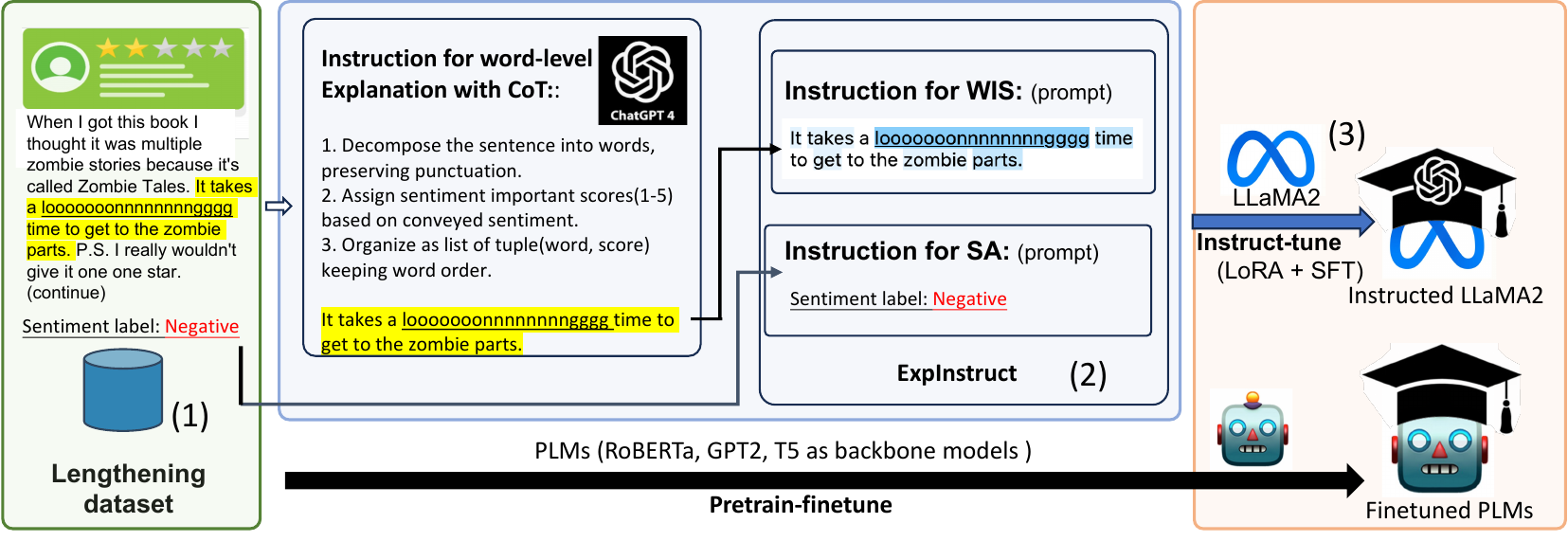}
\caption{An overview of our work for RLF. (1) We introduce \textbf{Lengthening} in Section \ref{sec:Lengthening_dataset}. (2) We propose the \textbf{ExpInstruct} framework and describe prompt details in Section \ref{sec:expInstruct}. (3) Experiments details are in Section \ref{sec:exp_setup}.}
\label{fig:instruct_llama2}
\end{figure*}

LMs consist of Pre-trained Language Models (PLMs) \cite{zhao2023survey} such as RoBERTa \cite{Yinhan2019RoBERTa}, T5 \cite{Raffel2019T5}, and GPT-2 \cite{radford2019GPT2}, and Large Language Models (LLMs), including GPT-4 \cite{openai2023gpt4} and LLaMA2 \cite{touvron2023llama2}. LMs exhibit impressive performance on many NLP tasks \cite{NEURIPS2020_1457c0d6, wang2023chatgptSentiment}. 
However, we know little about the boundaries of performance and explainability of LMs for RLF.
Such linguistic features, common in daily communication, have yet to be thoroughly investigated \cite{Go2009TwitterSC,Nguyen2017ADN,Abdul2017Emonet,Farman2019TransportationSentiment,Aljebreen_Meng_Dragut_2021}. The lack of a specialized dataset for RLF impedes us from evaluating and improving LMs to learn the nuanced communications in real-world and online social media content. 
This research gap raises two research questions: \textbf{1) Is RLF important for SA?} \textbf{2) Can LMs understand RLF?}
In this study, we aim to evaluate and improve the performance and explainability of PLMs and LLMs for RLF. 

The overview of our work is shown in Figure\ref{fig:instruct_llama2}.
To answer the first research question, we curate \textbf{Lengthening}, the first dataset focused on RLF for SA. Inspired by previous linguistic research of RLF \cite{KALMAN2014187, Gray_2020}, we design a pipeline for extracting RLF sentences and words from 4 public datasets and sample 850k instances. 
We conduct comprehensive experiments to compare zero-shot performance with 3 PLMs and 2 LLMs between RLF and w/o RLF groups. 
Our results reveal the sentiment-expressive value of RLF sentences with consistent higher performance. 
Importantly, we demonstrate the transferability of our fine-tuned LMs with \textbf{Lengthening} get document-level gain. Specifically, we observe an average improvement of a 7.6\%   in accuracy  and 4.5\% in F1 score (Table \ref{tab:document-gain}). 
These observations collectively show that RLF sentences can serve as key sentences for document-level SA tasks. In addition,
we highlight the potential of RLF for online social media content analysis, where short text and informal communications prevail. 

We study the second question and show that PLMs can reach better performances than zero-shot GPT-4 after fine-tuning on \textbf{Lengthening}, but lag behnd in explainability. 
We solve this issue with \textbf{ExpInstruct}, which can empower LLMs to reach the same level of performance and explainability as zero-shot GPT-4 with a small-size subset dataset.  
We analyze the data quality and explanation reliability with human evaluation. We further explore the effect of sample size and instruction strategy for ExpInstruct with ablation study.

In summary, our contributions are as follows:
\begin{itemize}
\item We call attention to RLF, an 
overlooked linguistic informal style. We show that RLF sentences can serve as signatures of document sentiment and have potential value for user generated content analysis.

\item We introduce \textbf{Lengthening}, a multi-domains dataset featuring RLFs with 850k samples grounding from 4 public datasets for SA tasks.

\item We propose a cost-effective approach \textbf{ExpInstruct}, which can improve the performance and explainability of open-sourced LLMs for RLF to the same level of zero-shot GPT-4. 

\item We quantify the explainability of PLMs and LLMs for RLF with a unified approach. Human evaluation demonstrates the reliability of this method.
\end{itemize}

\begin{table*}[ht]
\centering
\resizebox{\textwidth}{!}{
\begin{tabular}{lllcccccc}
\toprule
Domain  & Dataset & Original Tag & \#Document  & RLF Ratio(\%) & \#Samples & Label Distribution (1/0) & \#Unique RLF Word & \#Unique Root Word\\
\midrule
Books & Amazon Review & 1-5 stars & 51,312k &  5.41 & 255k & 92/8(\%) & 11,813 & 1,592\\
Electronics & Amazon Review & 1-5 stars & 20,994k  & 4.32 & 212k & 74/26(\%) & 11,367 & 1,541\\
Restaurants & Yelp  & 1-5 stars & 6,990k  & 11.28 & 315k & 74/26(\%) & 11,642 & 1,587\\
Social Media & Twitter & binary & 1,600k  & 13.36 & 44k & 46/54(\%) & 5,337 & 1,149\\
Hotels & TripAdvisor  & 1-5 stars & 879k   & 10.01 & 24k & 76/24(\%) & 4,558 & 1,250\\
\midrule
ALL & \textbf{Lengthening} & binary & 850k & 100 & 850k & 78/22(\%) & 19,610 & 1,677 \\
\bottomrule
\end{tabular}
}
\caption{Summary statistics for our dataset. \textbf{Lengthening} is the first large-scale dataset featuring RLF for SA task, grounded in 4 public datasets with an average of 5.8\% documents containing RLF. More details of the generation process of our dataset are described in Section \ref{sec:Lengthening_dataset}}
\label{tab:dataset_overview}
\end{table*}

\begin{table*}[ht]
\centering
\resizebox{1.0\textwidth}{!}{
\begin{tabular}{lllllcl}
\toprule
 Domain & Lengthening Style & POS & Lengthening Word & Root Word & Label & Example Sentence \\
 \midrule
Books & Punctuation & Noun & book!!!!! & book! & 1 & Do yourself a favour a read this book!!!!! \\
Electronics & Letter & Verb & loooove & love &1 & I loooove my new phone case. \\
Restaurants & Punctuation & Adj & amazing!!!!! & amazing! & 1 & We are from Seattle and this coffee is amazing!!!!!	 \\
Twitter & Letter & Adv & SOOOO & SO & 0 & SOOOO bummed i'm going to miss sam's party tonight. \\
Hotel & Punctuation & Noun & year............. & year... & 0 & I am looking to go back next year.............\\
\bottomrule
\end{tabular}
}
\caption{Samples from our \textbf{Lengthening} dataset.}
\label{tab:samples-with-pos-tag}
\end{table*}

\section{Related Work}
\textbf{Repetitive Lengthening Form (RLF)} 
The study by \cite{KALMAN2014187} shows that RLF is a written emulation of nonverbal spoken cues in user-generated content and proposes a method to reduce stretched words to their root words. 
\citet{Brody2011cool} highlights the importance of accurately interpreting RLF for SA and employing RLF to augment sentiment dictionary \cite{Construction_Sent_Dict}. \citet{schnoebelen2012emotions, Gray_2020} describe this nuanced linguistic feature as expressive words often used to emphasize or exaggerate the underlying sentiment intensity of the root word.
Previous research on RLF has primarily focused on linguistic and statistical analysis. A systematic study of the sentiment value of RLF in transform-based LMs is still lacking. 
We curate a large-scale dataset \textbf{Lengthening} with RLF for SA. We use the dataset to evaluate and improve the performance and explainability of the up-to-date PLMs and LLMs for RLF.

\noindent \textbf{Instruction Tuning} is a paradigm that fine-tunes language models on multiple tasks with instruction-input-output pairs to improve performance and generalize to unseen tasks \cite{Wei2021FinetunedLM, sanh2022multitask}. 
Emerging work explores instruction tuning to tasks such as text editing \cite{raheja-etal-2023-coedit}, information extraction \cite{lu-etal-2023-pivoine} and classification \cite{aly-etal-2023-automated}. 
We extend this line of work by instruct-tuning LLMs for SA with RLF. We focus on a novel task of predicting the document-level sentiment label using only one sentence.

\citet{honovich-etal-2023-unnatural, yin-etal-2023-dynosaur} use in-context learning strategies to prompt LLMs to automatically generate data for instruction tuning. \citet{lampinen-etal-2022-language, zhang-etal-2023-instructsafety, zhou-etal-2023-flame} instruct LLMs with emphasis on explainability.
Inspired by those works, we prompt GPT-4 with Chain of Thought (CoT) \cite{wei2023chainofthought} to generate word importance scores for RLF sentences. Using these scores along with the ground truth document-level sentiment labels, we automatically generate prompts for explainable instruction tuning of LLMs.

\noindent \textbf{Model Explanation} 
The importance of input features has been extensively explored in recent years. Studies such as \cite{Marco2016LIME,li2017understanding,zhu2023promptbench,ray2022MachineReading,atanasova-etal-2020-diagnostic} have employed saliency-based methods for PLMs which are based on changes in loss or gradient metrics. 
The approach does not apply to prompt-interaction LLMs. Furthermore, the loss or gradient values are not always available for closed-source LLMs (e.g., GPT-4 and Claude).
Prompting methods with CoT are proposed for feature importance analysis with LLMs \cite{zhong2023ChatgptBert, wang2023chatgptSentiment}.
However, the difference between saliency and prompt based methods creates a barrier to comparing explainability results between PLMs and LLMs.
We overcome that challenge by proposing a unified approach for evaluating the explainability of LMs in handling RLF
(Section \ref{sec:unified_exp}).

\section{The Lengthening Dataset}
\label{sec:Lengthening_dataset}
We introduce the \textbf{Lengthening} dataset in this section.
We need a few definitions first:
RLF sentence is a sentence with one or more RLF words, RLF document is a document (e.g., user review) with at least one RLF sentence.
We present an overview of data statistics for our dataset and present in Table\ref{tab:dataset_overview}. 
More detailed examples from our dataset can be found in Table \ref{tab:samples-with-pos-tag}.

\subsection{Data Source}
To comprehensively evaluate the usage of RLF in social media platforms and online user reviews,
we select 4 public datasets covering 5 distinct domains: \textbf{Books \& Electronics} from Amazon Reviews \cite{ni2019justifying}; \textbf{Restaurant Reviews} from Yelp \cite{yelpdataset} data from Feb 16, 2021; \textbf{Hotel Reviews} from TripAdvisor \cite{li2020hotel}; \textbf{Twitter} dataset with general social posts \cite{Go2009TwitterSC}. 
All user reviews (documents) are categorized based on their star ratings: 1-2 stars as negative, 4-5 stars as positive, with 3-star reviews being excluded as they are assumed to be neutral.

\subsection{Generation of Lengthening}

We describe our pipeline for extracting RLF sentences and words from documents with three steps. 
We give additional details about dataset generation techniques and algorithms in Appendix \ref{sec:dataset_details}. 

\noindent \textbf{Identification of Potential RLF Documents}
We design a regular expression, termed RLFsearch, to identify potential RLF documents by matching repeated letters and repeated punctuation patterns \cite{Gray_2020}. Documents that return a positive result from RLFsearch are retained for the next step.

\noindent \textbf{Potential RLF Sentences Extraction}
We segment potential RLF documents into sentences.
Sentences containing fewer than five words are merged with the preceding sentences to maintain the overall sentiment polarity \cite{dragut-etal-2012-Polarity,Eduard_Polarity_Domain}.
Each sentence is re-evaluated using the RLFsearch regex to extract potential RLF sentences.

\noindent \textbf{RLF Extraction}
We further split potential RLF sentences into individual words and apply the RLFsearch regex at word-level. This step excludes numbers, URLs, words beginning with '@', and monetary amounts. Words that pass this regex filter are recognized as RLF words. Sentences that contain one or more RLF words are extracted as RLF sentences. Correspondingly, documents with one or more RLF sentences are extracted as RLF documents. 
For each RLF sentence, a sentence without any RLFs (w/o RLF) from the same document is randomly chosen for zero-shot comparison (we only consider documents containing two or more sentences). 
In addition, we find the root word 
of each RLF word with the algorithm proposed in \cite{KALMAN2014187} based on American English. 
For example, RLF words like `loooove' and `loooovvve' have the same root word `love'.


\section{Method}
We introduce ExpInstruct, a two-stage instruction tuning framework aimed to improve both performance and explainability of LLMs for RLF. 
ExpInstruct first prompts GPT-4 for WIS scores and then finetunes LLMs with instructions for WIS and SA.
We further propose a unified approach to evaluate the comprehension level of LMs for RLF.

Formally, consider \( (x, rlf, y) \) to be a single tuple in our dataset \( D \), where \( x \) is an RLF sentence with an RLF word \(rlf\)  and \(y\) is the document-level sentiment label. We denote a transformer-based model by \( f \).

\subsection{ExpInstruct}
\label{sec:expInstruct}
\begin{figure}[t]
\centering
\includegraphics[width=\columnwidth]{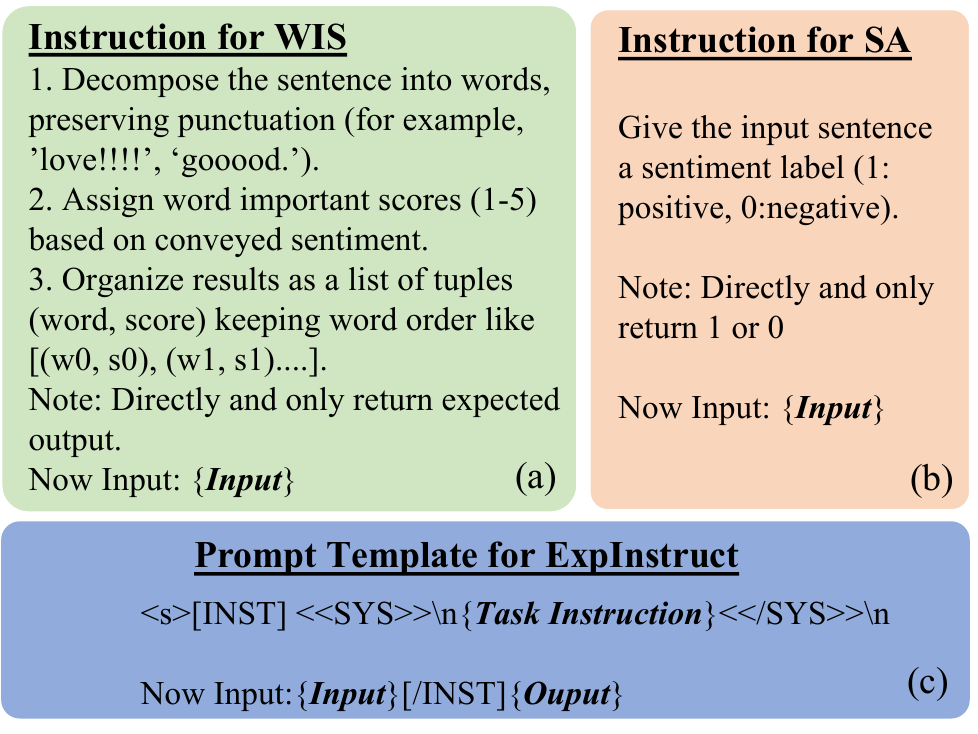}
\caption{Prompt Design and Template for ExpInstruct. (a) Prompt with CoT for word-level explainability. 
(b) Simple Prompt for SA. (c) Prompt Template for Instruction tuning}
\label{fig:prompt_WIS}
\end{figure}

\textbf{Prompt Design for Explanation} We prompt GPT-4 with CoT to generate Word Importance Scores (WIS) to reflect word-level understanding of input sentence $x$ as shown in Figure \ref{fig:prompt_WIS}(a). 
The CoT consists of 3 sequential reasoning steps. 1) Sentence Decomposition: Segment sentences into words and keep punctuation marks with a few-shot strategy. 2) Word Importance Scoring: Assign sentiment importance scores (1-5) to each word, which can reflect LLMs' understanding of word-level sentiment. 3) Structured Output: We specify the structured output format for subsequent analysis. 
 
\noindent \textbf{Instruction Template}
ExpInstruct has two tasks with the same prompt template as shown in Figure \ref{fig:prompt_WIS}(c). This is achieved by adding three placeholders: \{Task Instruction\}, \{Input\}, and \{Output\}.
The first task is \textit{Instruction for WIS}: The Task Instruction is shown in Figure \ref{fig:prompt_WIS}(a), with the Input as RLF sentence $x$ and the Output is the structured output generated by GPT-4 for WIS.
The second task is \textit{Instruction for SA}: The Task Instruction is shown in Figure \ref{fig:prompt_WIS}(b); the Input is the RLF sentence $x$ and the Output is the document-level sentiment label $y$.

\subsection{A Unified Approach to Evaluate Explainability}
\label{sec:unified_exp}
In this section, we propose a unified approach to evaluate the explainability of PLMs and LLMs with the help of WIS. 
Our approach consists of two steps: 1) Generate WIS from LLMs and PLMs, and 2) Quantify explainability across models with normalization.

\noindent \textbf{Generate WIS}
For LLMs, we use a prompt-based method to generate WIS \cite{zhong2023ChatgptBert, wang2023chatgptSentiment} with the instruction shown in Figure  \ref{fig:prompt_WIS}(a). This method requires only one-time inference for each input sentence $x$ and is label-free:
\begin{equation}
\label{eq:llm_prompt}
\text{WIS} = f(x, \text{Instruction for WIS}) 
\end{equation}

For PLMs, we choose a saliency-based method to generate WIS. Specifically, we use the occlusion-based method \cite{ray2022MachineReading, zhu2023promptbench} because it's intuitive and applicable to various PLMs. This method involves sequentially removing one word \( w_i \)
to observe the absolute change in the loss value, serving as an indicator of the word-level significance:
\begin{equation}
\label{eq:occ_score}
\text{WIS}[w_i] = |L[f(x), y] - L[f(x-w_i), y]|
\end{equation}
where \(f(x)\) represents the output logit value from the transformer-based model $f$ given the input $x$, $L$ is the cross-entropy loss function, and \( x - w_i \) is the sentence after the removal of the word \( w_i \) from the input sentence.

\begin{figure}[ht]
\centering
\resizebox{0.4\textwidth}{!}{
\includegraphics[width=1.0\textwidth]{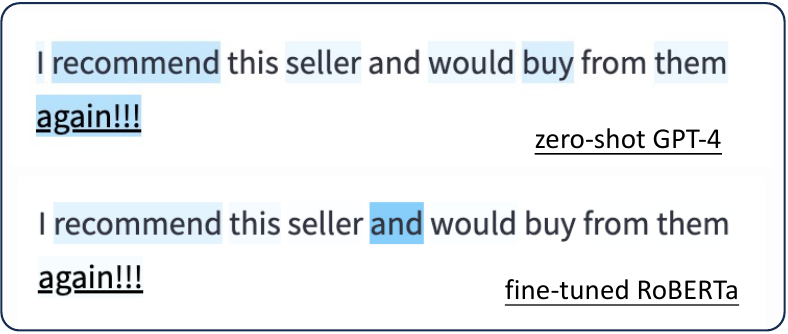}
}
\caption{Comparing normalized WIS for an RLF sentence from zero-shot GPT-4 and fine-tuned RoBERTa.}
\label{fig:WIS_example}
\end{figure}

\noindent \textbf{Quantify Explainability}
To eliminate the barrier caused by differing relative WIS values across models, we apply min-max followed by \( L_1 \) normalization to WIS and denote the normalized WIS as \(\text{WIS}_{\text{norm}}\). 
We visualize the normalized WIS from zero-shot GPT-4 and fine-tuned RoBERTa with a sample sentence in Figure \ref{fig:WIS_example}.

The overall explainability score \( S_{exp} \) for a given dataset $D$ can be quantified as:
\begin{equation}
S_{exp}  =  \frac{1}{|D|} \sum\limits_{(x,rlf,y) \in D} \  \text{WIS}_{\text{norm}}\left[ j_{(w_j = rlf)} \right]
\label{eq:all_occ_ev}
\end{equation}
where $j$ refers to the word index where $w_j = rlf$ in the input sentence $x$.

We take an average of the \(\text{WIS}_{\text{norm}}\) of RLF  for all $(x, rlf, y)$ pairs to compute the explainability score $S_{exp}$ for the target model $f$ on dataset $D$.
A higher $S_{exp}$ score indicates that the model pays more attention to RLF words for the SA task, which reflects a better understanding of the expressive value of RLF.
This metric enables us to automatically quantify the explainability of LMs for RLF. It can be used as a complement to the qualitative method which relies on case studies and human verification \cite{ray2022MachineReading, zhu2023promptbench}.


\section{Experimental Setup}
\label{sec:exp_setup}
In this section, we introduce baseline models, experimental design and implementation details.

\noindent \textbf{Baseline Models}
To explores the boundaries of performance and explainability for SOTA PLMs and LLMs for RLF,
we choose 3 fine-tuned 3 PLMs for SA task with backbone models as RoBERTa (Large), GPT-2 (Medium) and T5 (Base). These models have comparable parameter scales and represent encoder-only, decoder-only, and encoder-decoder architectures.  
For LLMs, we use GPT-4 via the OpenAI API. 
Moreover, we select LLaMA2 (13B-chat-hf) for instruct-tuning because it's open-sourced with auto-regressive architecture like GPT-4. And the scalable parameter size of LLaMA2 (13B) allows inference and fine-tuning with LoRA \cite{hu2021lora} on a single GPU. 
More details of these baseline models and API usage are provided in the Appendix \ref{sec:model_par}

\noindent \textbf{Implementation}
\label{sec:Implementation}
We fine-tune 3 PLMs on the entire Lengthening dataset. The batch sizes are set to 64 for RoBERTa and T5, and 32 for GPT-2 to optimize GPU memory usage.
We sampled 3,000 instances from Lengthening as a subset dataset for experiments with LLMs with stratified sampling method based on domains, where we used 1.6k instances for training and others for validation and testing.
This sample size aligns with previous studies \cite{wang2023chatgptSentiment,zhang2023instructfingpt,deng2022llmsReddit}. More details of data split see Appendix \ref{sec:dataset_split}.

We instruct-tune LLaMA2 with the our ExpInstruct framework with  LoRA and Supervised fine-tuning (SFT) \cite{vonwerra2022trl} to lower computational costs. We set the temperature at 0.2 for all LLMs.
All models are trained for a maximum of 5 epochs with k-fold cross-validation strategy (k = 3) to identify the best checkpoints. 
In each iteration one fold serves as the test set while the remaining data is split into train/val with 4/1 ratio.
It takes one week to conduct all experiments with a single RTX A6000 GPU with 50 GB of memory.

\begin{table}[ht]
\centering
\resizebox{\columnwidth}{!}{
\begin{tabular}{lcccc}
\toprule
Metric & \multicolumn{2}{c}{Acc(\%)} & \multicolumn{2}{c}{F1(\%)}  \\
 \cmidrule(lr){2-3} \cmidrule(lr){4-5}  
 Backbone Model & RLF & w/o RLF & RLF & w/o RLF \\
 \midrule
RoBERTa (Large) & \underline{85.94} \footnotesize{$\pm$ 0.03} & \underline{84.40} \footnotesize{$\pm$ 0.04} & \textbf{90.67} \footnotesize{$\pm$ 0.04} & \underline{89.56} \footnotesize{$\pm$ 0.03} \\
GPT-2 (Medium) & 79.56 \footnotesize{$\pm$ 0.07} & 77.56 \footnotesize{$\pm$ 0.08} & 85.76 \footnotesize{$\pm$ 0.08} & 84.22 \footnotesize{$\pm$ 0.06} \\
T5 (Base) & 83.22 \footnotesize{$\pm$ 0.06} & 81.93 \footnotesize{$\pm$ 0.04} & 88.71 \footnotesize{$\pm$ 0.07} & 87.85 \footnotesize{$\pm$ 0.03} \\
LLaMA2 (13B) & 76.25 \footnotesize{$\pm$ 1.76} & 70.41 \footnotesize{$\pm$ 0.57} & 84.30 \footnotesize{$\pm$ 1.28} & 82.64 \footnotesize{$\pm$ 0.39} \\
GPT-4 & \textbf{86.26} \footnotesize{$\pm$ 0.93} & \textbf{86.20} \footnotesize{$\pm$ 1.45} & \underline{90.12} \footnotesize{$\pm$ 0.75} & \textbf{90.08} \footnotesize{$\pm$ 1.14} \\
\bottomrule
\end{tabular}
}
\caption{Overall zero-shot accuracy and F1 score for sentences with RLF and without RLF words (w/o RLF). Bold denotes the best results in a column 
and underline highlights the second best results.
$\pm$ indicates standard deviation score.} 
\label{tab:zero-shot-overall}
\end{table}

\begin{figure*}[ht] 
   \begin{subfigure}{0.31\textwidth}
       \includegraphics[width=\linewidth]{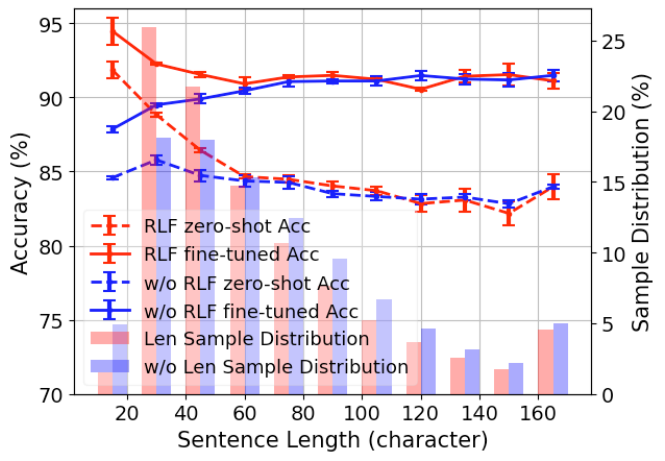}
       \caption{RoBERTa (Large)}
       \label{fig:Roberta-large}
   \end{subfigure}
\hfill 
   \begin{subfigure}{0.31\textwidth}
       \includegraphics[width=\linewidth]{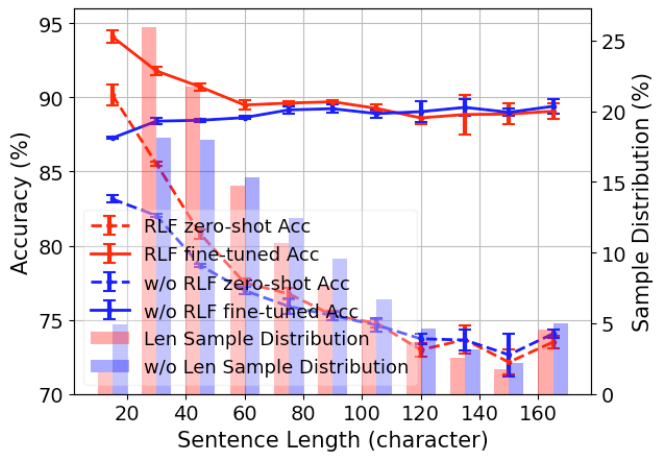}
       \caption{GPT-2 (Medium)}
       \label{fig:GPT2-medium}
   \end{subfigure}
\hfill 
   \begin{subfigure}{0.31\textwidth}
       \includegraphics[width=\linewidth]{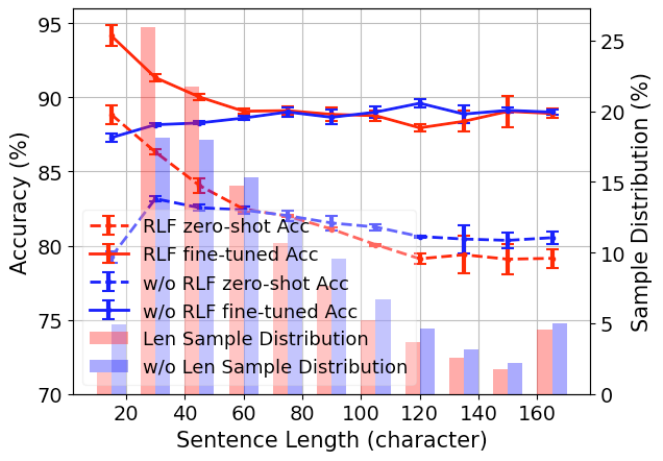}
       \caption{T5 (Base)}
       \label{fig:T5-small}
   \end{subfigure}

   \caption{Comparison of accuracy between the RLF and w/o RLF groups using zero-shot and fine-tuned models by sentence length. The lines represent average values, and the error bar indicate the standard deviation for each length group across 3 runs. Both results across 3 fine-tuned models show a convergence between the RLF and w/o RLF groups when the sentence character length is around 80.}
   \label{fig:accuracy-sentence-length}
\end{figure*}

\section{Results}

\textbf{Is RLF Important for Sentiment Analysis?}
We approach this question from a new angle: predicting the document-level sentiment label with a single sentence.
Specifically, we compare the performance of two groups: sentences with RLF and those without (w/o RLF).
Evaluations are conducted for zero-shot and fine-tuned models with Accuracy (Acc) and macro F1 score (F1) as performance metrics.

Firstly, we present the zero-shot results in Table \ref{tab:zero-shot-overall}. 
The RLF group consistently achieves better performance than the w/o RLF group in both Accuracy and F1 score across all models.
These results indicate that sentences with RLF can serve as key signatures for document-level sentiment.
To our knowledge, this study is the first to empirically demonstrate the sentiment-expressive value of RLF through comprehensive experiments with various PLMs and LLMs.

\begin{table*}[ht]
\centering
\resizebox{\textwidth}{!}{
\begin{tabular}{lcccccccccccc}
\toprule
Backbone Model &  \multicolumn{4}{c}{RoBERTa (Large)} & \multicolumn{4}{c}{GPT-2 (Medium)} & \multicolumn{4}{c}{T5 (Base)}  \\
 \cmidrule(lr){2-5} \cmidrule(lr){6-9} \cmidrule(lr){10-13} 
Metric &  \multicolumn{2}{c}{Acc(\%)} & \multicolumn{2}{c}{F1(\%)} & \multicolumn{2}{c}{Acc(\%)} & \multicolumn{2}{c}{F1(\%)} & \multicolumn{2}{c}{Acc(\%)} & \multicolumn{2}{c}{F1(\%)}  \\
  \cmidrule(lr){2-3} \cmidrule(lr){4-5} \cmidrule(lr){6-7} \cmidrule(lr){8-9} \cmidrule(lr){10-11}    \cmidrule(lr){12-13}
 Domain / Group & RLF & w/o RLF & RLF & w/o RLF  & RLF & w/o RLF & RLF & w/o RLF & RLF & w/o RLF & RLF & w/o RLF  \\
 \midrule
Books & \textbf{87.36} \footnotesize{$\pm$ 0.17} & 86.90 \footnotesize{$\pm$ 0.16} & \textbf{92.76} \footnotesize{$\pm$ 0.10} & 92.46 \footnotesize{$\pm$ 0.10} & 80.65 \footnotesize{$\pm$ 0.07} & 81.21 \footnotesize{$\pm$ 0.18} & 88.45 \footnotesize{$\pm$ 0.06} & 88.84 \footnotesize{$\pm$ 0.11} & 86.09 \footnotesize{$\pm$ 0.17} & \underline{87.07} \footnotesize{$\pm$ 0.08} & 92.01 \footnotesize{$\pm$ 0.10} & \underline{92.63} \footnotesize{$\pm$ 0.05} \\
Restaurants & \textbf{88.46} \footnotesize{$\pm$ 0.07} & \underline{85.93} \footnotesize{$\pm$ 0.09} & \textbf{92.07} \footnotesize{$\pm$ 0.07} & \underline{90.24} \footnotesize{$\pm$ 0.06} & 82.67 \footnotesize{$\pm$ 0.20} & 79.04 \footnotesize{$\pm$ 0.02} & 87.60 \footnotesize{$\pm$ 0.18} & 84.70 \footnotesize{$\pm$ 0.02} & 85.26 \footnotesize{$\pm$ 0.23} & 82.79 \footnotesize{$\pm$ 0.14} & 89.70 \footnotesize{$\pm$ 0.19} & 87.95 \footnotesize{$\pm$ 0.09} \\
Electronics & \textbf{84.56} \footnotesize{$\pm$ 0.08} & \underline{82.54} \footnotesize{$\pm$ 0.19} & \textbf{88.86} \footnotesize{$\pm$ 0.10} & \underline{87.27} \footnotesize{$\pm$ 0.16} & 76.35 \footnotesize{$\pm$ 0.09} & 73.28 \footnotesize{$\pm$ 0.17} & 81.72 \footnotesize{$\pm$ 0.16} & 78.99 \footnotesize{$\pm$ 0.21} & 80.53 \footnotesize{$\pm$ 0.12} & 77.84 \footnotesize{$\pm$ 0.05} & 85.65 \footnotesize{$\pm$ 0.02} & 83.54 \footnotesize{$\pm$ 0.09} \\
Twitter & 66.30 \footnotesize{$\pm$ 0.36} & \textbf{66.87} \footnotesize{$\pm$ 0.40} & \textbf{68.64} \footnotesize{$\pm$ 0.35} & \underline{68.41} \footnotesize{$\pm$ 0.43} & \underline{66.84} \footnotesize{$\pm$ 0.13} & 65.71 \footnotesize{$\pm$ 0.57} & 66.11 \footnotesize{$\pm$ 0.14} & 65.53 \footnotesize{$\pm$ 0.98} & 65.55 \footnotesize{$\pm$ 0.35} & 64.56 \footnotesize{$\pm$ 0.05} & 66.36 \footnotesize{$\pm$ 0.50} & 65.22 \footnotesize{$\pm$ 0.19} \\
Hotel & \underline{84.07} \footnotesize{$\pm$ 0.18} & \textbf{84.37} \footnotesize{$\pm$ 0.60} & \underline{89.11} \footnotesize{$\pm$ 0.16} & \textbf{89.32} \footnotesize{$\pm$ 0.43} & 76.55 \footnotesize{$\pm$ 0.20} & 76.27 \footnotesize{$\pm$ 0.49} & 83.14 \footnotesize{$\pm$ 0.06} & 82.97 \footnotesize{$\pm$ 0.39} & 80.19 \footnotesize{$\pm$ 0.30} & 81.43 \footnotesize{$\pm$ 0.10} & 86.19 \footnotesize{$\pm$ 0.23} & 87.24 \footnotesize{$\pm$ 0.02} \\
\bottomrule
\end{tabular}
}
\caption{Zero-shot accuracy and F1 score for sentences with/without RLF words  in different domains. We report RoBERTa (Large), GPT-2 (Medium) and T5 (Base) because limited test samples for GPT-4 and LLaMA2-13B. Bold and underline indicate best and second best results. 
$\pm$ indicates standard deviation score.
}
\label{tab:zero-shot-domain}
\end{table*}

\begin{table*}[ht]
\centering
\resizebox{\textwidth}{!}{
\begin{tabular}{lccccccccc}

\toprule
 Lengthening Style & \multicolumn{3}{c}{Punctuation Repetitive} & \multicolumn{3}{c}{Letter Repetitive} & \multicolumn{3}{c}{Overall}\\
 \cmidrule(lr){2-4} \cmidrule(lr){5-7} \cmidrule(lr){8-10} 
Backbone Model & $S_{exp}$ &  Acc(\%) & F1(\%) & $S_{exp}$  & Acc(\%) & F1(\%) & $S_{exp}$  & Acc(\%) & F1(\%) \\
\midrule

\textit{\textbf{Zero-shot}} \\

RoBERTa (Large) & 0.24 \footnotesize{$\pm$ .001} & \underline{86.65} \footnotesize{$\pm$ 0.01} & \underline{91.19} \footnotesize{$\pm$ 0.01} & 0.10 \footnotesize{$\pm$ .001} & \textbf{83.61} \footnotesize{$\pm$ 0.16} & \textbf{88.93} \footnotesize{$\pm$ 0.17} & 0.21 \footnotesize{$\pm$ .001} & \underline{85.94} \footnotesize{$\pm$ 0.03} & \textbf{90.67} \footnotesize{$\pm$ 0.04}\\
GPT-2 (Medium) & \textbf{0.49} \footnotesize{$\pm$ .001} & 80.55 \footnotesize{$\pm$ 0.11} & 86.57 \footnotesize{$\pm$ 0.08} & 0.07 \footnotesize{$\pm$ .001} & 76.34 \footnotesize{$\pm$ 0.11} & 82.98 \footnotesize{$\pm$ 0.15} & \textbf{0.39} \footnotesize{$\pm$ .002} & 79.56 \footnotesize{$\pm$ 0.07} & 85.76 \footnotesize{$\pm$ 0.08}\\
T5 (Base) & 0.20 \footnotesize{$\pm$ .001} & 83.97 \footnotesize{$\pm$ 0.04} & 89.30 \footnotesize{$\pm$ 0.04} & 0.11 \footnotesize{$\pm$ .001} & 80.75 \footnotesize{$\pm$ 0.10} & \underline{86.74} \footnotesize{$\pm$ 0.16} & 0.18 \footnotesize{$\pm$ .001} & 83.22 \footnotesize{$\pm$ 0.06} & 88.71 \footnotesize{$\pm$ 0.07}\\
LLaMA2 (13B) & 0.18 \footnotesize{$\pm$ .004} & 77.25 \footnotesize{$\pm$ 2.62} & 85.23 \footnotesize{$\pm$ 1.89} & \underline{0.25} \footnotesize{$\pm$ .003} & 73.54 \footnotesize{$\pm$ 3.86} & 81.46 \footnotesize{$\pm$ 2.88} & 0.20 \footnotesize{$\pm$ .004} & 76.25 \footnotesize{$\pm$ 1.76} & 84.30 \footnotesize{$\pm$ 1.28} \\
GPT-4  & \underline{0.39} \footnotesize{$\pm$ .005} & \textbf{87.69} \footnotesize{$\pm$ 1.66} & \textbf{91.32} \footnotesize{$\pm$ 1.35} & \textbf{0.34} \footnotesize{$\pm$ .004} & \underline{82.36} \footnotesize{$\pm$ 4.00} & 86.40 \footnotesize{$\pm$ 3.39} & \underline{0.38} \footnotesize{$\pm$ .005} & \textbf{86.26} \footnotesize{$\pm$ 0.93} & \underline{90.12} \footnotesize{$\pm$ 0.75} \\
\midrule

\textit{\textbf{Fine-tuned}} \\

RoBERTa (Large) & 0.24 \footnotesize{$\pm$ .012} & \textbf{91.97} \footnotesize{$\pm$ 0.02} & \textbf{94.85} \footnotesize{$\pm$ 0.02} & \underline{0.14} \footnotesize{$\pm$ .005} & \textbf{90.33} \footnotesize{$\pm$ 0.25} & \textbf{93.71} \footnotesize{$\pm$ 0.21} & 0.22 \footnotesize{$\pm$ .010} & \textbf{91.59} \footnotesize{$\pm$ 0.08} & \textbf{94.59} \footnotesize{$\pm$ 0.07}\\
GPT-2 (Medium) & \underline{0.34}   \footnotesize{$\pm$ .006} & \underline{90.80} \footnotesize{$\pm$ 0.09} & \underline{94.13} \footnotesize{$\pm$ 0.07} & 0.12 \footnotesize{$\pm$ .001} & \underline{88.92} \footnotesize{$\pm$ 0.38} & \underline{92.82} \footnotesize{$\pm$ 0.29} & \underline{0.29} \footnotesize{$\pm$ .005} & \underline{90.36} \footnotesize{$\pm$ 0.16} & \underline{93.83} \footnotesize{$\pm$ 0.12} \\
T5 (Base) & 0.25  \footnotesize{$\pm$ .002}  & 90.36 \footnotesize{$\pm$ 0.07} & 93.88 \footnotesize{$\pm$ 0.05} & 0.13 \footnotesize{$\pm$ .001} & 88.17 \footnotesize{$\pm$ 0.19} & 92.34 \footnotesize{$\pm$ 0.15} & 0.23 \footnotesize{$\pm$ .002} & 89.85 \footnotesize{$\pm$ 0.08} & 93.52 \footnotesize{$\pm$ 0.05} \\
\textbf{ExpInstruct} & \textbf{0.37} \footnotesize{$\pm$ .019} & 88.46 \footnotesize{$\pm$ 0.83} & 92.05 \footnotesize{$\pm$ 0.65} & \textbf{0.30} \footnotesize{$\pm$ .021} & 83.79 \footnotesize{$\pm$ 2.74} & 87.58 \footnotesize{$\pm$ 2.53} & \textbf{0.35} \footnotesize{$\pm$ .019} & 87.20 \footnotesize{$\pm$ 0.69} & 90.96 \footnotesize{$\pm$ 0.51} \\

\bottomrule
\end{tabular}
}
\caption{Comparision performances between zero-shot and finetuned models with two lengthening styles. In each column, the \textbf{best} result is highlighted in bold, and the \underline{second best} result is underlined.}
\label{tab:len-style-exp-metrics}
\end{table*}

Furthermore, we conduct a domain-wise analysis of zero-shot performance as shown in Table \ref{tab:zero-shot-domain}. 
Due to sample size limitations for the subset dataset, this analysis is focused on PLMs. 
The superiority of the RLF sentences in expressing sentiment is consistently observed across diverse domains and models, highlighting the robustness and generalizability of the sentiment-expressive value of RLF.

In Figure \ref{fig:accuracy-sentence-length}, we present the accuracy of zero-shot and fine-tuned PLMs in relation to sentence length. 
We observe performance improvements in all PLMs for both the RLF and w/o RLF groups after fine-tuning on our \textbf{Lengthening} dataset.
Superisingly, the RLF group demonstrates significantly better performance than the w/o RLF group when sentence lengths are within 80 characters, with more than 70\% of sentences within this range.
This performance gap is evident in both zero-shot and fine-tuned models.
These findings highlight the critical importance of focusing on RLF in social media content, where short and informal expressions prevail.

\noindent \textbf{Can LMs Understand RLF?} 
To answer this question, we compare the performance and explainability of zero-shot and fine-tuned models for RLF sentences with Acc, F1 and $S_{exp}$ as evaluation metrics.
We present the results for two RLF styles in Table\ref{tab:len-style-exp-metrics}.
Our results show that GPT-4 has the best performance and explainability of RLF among zero-shot models. 
Although zero-shot GPT-2 (Medium) has the highest $S_{exp}$ score, its low Acc and F1 scores suggest an insufficient understanding of RLF. This is further supported by the drop of $S_{exp}$ score after fine-tuning GPT-2 (Medium).
We observe that all fine-tuned PLMs achieve better Acc and F1 scores compared to zero-shot GPT-4. However, their $S_{exp}$ scores are still lower than zero-shot GPT-4 with a significant gap, suggesting that the fine-tuned PLMs may lack sufficient understanding of RLF. 
This 
appears to be another instance of LMs  being ``right for the wrong reasons''\cite{mccoy-etal-2019-right}.

Interestingly, our ExpInstruct with LLaMA2 achieves the same level of performance and explainability as zero-shot GPT-4 with only 1,600 instruction samples.
This finding highlights the effectiveness of ExpInstruct, a cost-effective approach that requires only limited samples for instruction tuning to enhance open-source LLMs as alternatives to GPT-4.

\section{Analysis}
In this section, we first conduct human evaluation for data quality and explanation reliability.
Additionally, we verify the benefits gained from \textbf{Lengthening} are transferable to document-level SA.
Moreover, ablation studies are conducted to explore the effects of training sample size and instruct strategy. 

\noindent \textbf{Data Quality} 
The final sentiment label for each sentence was determined by majority vote among 3 annotators. 
We use Krippendorff's Alpha score \cite{krippendorff2018content} for inter-rater agreement (IAA) score and obtained a score of 0.86, indicating that the annotated data is reliable.
We report human performance on our sentence to document SA task with Acc / F1 score as 90.01 / 92.52\% for the RLF group and 85.50 / 89.04\% for the w/o RLF group.
The superior scores of the RLF group support our main conclusion that RLF sentences can serve as key sentence for document-level SA.
This result shows that zero-shot GPT-4 is close to human-level performance in our task.
We further present the confusion matrix for sample distribution in Table \ref{tab:confusion_matrices}.

\begin{table}[ht]
\centering
\resizebox{0.3\textwidth}{!}{
\begin{tabular}{lcccc}
\toprule
& PP & PN & NP & NN \\
\hline
RLF & 127 & 9 & 11 & 53 \\
w/o RLF & 120 & 16 & 13 & 51 \\
\bottomrule
\end{tabular}
}
\caption{Confusion matrices for sample distribution. We categorized data with combinations of document and sentence labels (e.g., PP represents Positive document with Positive sentence).}
\label{tab:confusion_matrices}
\end{table}

\noindent \textbf{Explanation Reliability}
We ask the annotators to evaluate the reliability of the WIS generated by the zero-shot GPT-4 and the four fine-tuned LMs (RoBERTa, GPT-2, T5, and LLaMA2) with criteria of 1: Agree, 0: Disagree.
The final reliability score for each sample was determined by averaging scores among annotators. 
We report a moderate overall IAA score of 0.44 and show detailed result in Table \ref{tab:WIS_he}.
This result supports our conclusion that fine-tuned PLMs still have a gap in understanding RLF compared to zero-shot GPT-4.  
Furthermore, the correlation coefficient between the reliability score and $S_{exp}$ is 0.91, showing the validity of our unified approach for explainability evaluation proposed in Section \ref{sec:unified_exp}. 

\begin{table}[ht]
\centering
\resizebox{0.45\textwidth}{!}{
\begin{tabular}{lccc}
\toprule
Backbone Model & IAA & Reliability & $S_{exp}$ \\
\hline
RoBERTa (Large) & 0.41 & 0.57 \footnotesize{$\pm$ .159} & 0.22 \footnotesize{$\pm$ .010} \\
GPT-2 (Medium) & 0.44  & 0.59 \footnotesize{$\pm$ .110} & 0.29 \footnotesize{$\pm$ .005} \\
T5 (Base) & 0.42 & 0.48 \footnotesize{$\pm$ .093} & 0.23 \footnotesize{$\pm$ .002} \\
GPT-4 & 0.43 & 0.79 \footnotesize{$\pm$ .055} & 0.38 \footnotesize{$\pm$ .005} \\
\textbf{ExpInstruct} & 0.34 & 0.67 \footnotesize{$\pm$ .105} & 0.35 \footnotesize{$\pm$ .019} \\
\bottomrule
\end{tabular}
}
\caption{Detail results of Explanation Reliability for zero-shot GPT-4 and the 4 fine-tuned models
(RoBERTa, GPT-2, T5, and ExpInstruct). The correlation coefficient between the reliability score and $S_{exp}$ is 0.91.}
\label{tab:WIS_he}
\end{table}

\begin{table}[ht]
\centering
\resizebox{0.45\textwidth}{!}{%
\begin{tabular}{lcccc}
\toprule
Metric & \multicolumn{2}{c}{Acc(\%)} & \multicolumn{2}{c}{F1(\%)}  \\
 \cmidrule(lr){2-3} \cmidrule(lr){4-5}  
Backbone Model & Zero-shot & Fine-tuned & Zero-shot & Fine-tuned \\
 \midrule
RoBERTa (Large) & 93.99 \footnotesize{$\pm$ 0.13} & 95.99 \footnotesize{$\pm$ 0.51} $\uparrow$ & 95.82 \footnotesize{$\pm$ 0.13} & 97.18 \footnotesize{$\pm$ 0.36} $\uparrow$ \\
GPT-2 (Medium) & 89.79 \footnotesize{$\pm$ 0.38} & 94.14 \footnotesize{$\pm$ 0.16} $\uparrow$ & 92.72 \footnotesize{$\pm$ 0.35} & 95.94 \footnotesize{$\pm$ 0.11} $\uparrow$ \\
T5 (Base) & 92.61 \footnotesize{$\pm$ 0.64} & 93.37 \footnotesize{$\pm$ 0.06} $\uparrow$ & 94.84 \footnotesize{$\pm$ 0.47} & 95.38 \footnotesize{$\pm$ 0.04} $\uparrow$ \\
LLaMA2 (13B) & 71.06 \footnotesize{$\pm$ 0.42} & 94.18 \footnotesize{$\pm$ 0.63} $\uparrow$ & 83.04 \footnotesize{$\pm$ 0.34} & 95.96 \footnotesize{$\pm$ 0.37} $\uparrow$ \\
\bottomrule
\end{tabular}%
}
\caption{Document-level gains for LMs fine-tuned with \textbf{Lengthening}. We observe average improvement in accuracy as 7.6\% and F1 score as 4.5\% among the models.} 
\label{tab:document-gain}
\end{table}

\begin{table}[ht]
\centering
\resizebox{0.45\textwidth}{!}{
\begin{tabular}{lccc}
\toprule
 & $S_{exp}$  & Acc(\%) & F1(\%) \\
  \midrule
\textbf{\textit{\# Training Samples}} \\
0  &  0.20  \footnotesize{$\pm$ .004} & 76.25  \footnotesize{$\pm$ 1.76} & 84.30 \footnotesize{$\pm$ 1.28} \\
500  &  0.33 \footnotesize{$\pm$ .005} & 82.09  \footnotesize{$\pm$ 0.89} & 86.94 \footnotesize{$\pm$ 0.35} \\
1,000  &  0.35 \footnotesize{$\pm$ .015} & 85.36 \footnotesize{$\pm$ 0.18} & 89.58 \footnotesize{$\pm$ 0.26} \\
\textbf{ExpInstruct (1,600)}  & \textbf{0.35} \footnotesize{$\pm$ .019} & \textbf{87.20} \footnotesize{$\pm$ 0.69} & \textbf{90.96} \footnotesize{$\pm$ 0.51} \\
\midrule
\textit{\textbf{Instruction Strategy}} \\
Instruction with SA &  0.15 \footnotesize{$\pm$ .003} & 86.61 \footnotesize{$\pm$ 0.77} & 90.57 \footnotesize{$\pm$ 0.70} \\
Instruction with WIS &  0.35 \footnotesize{$\pm$ .005} & 72.60 \footnotesize{$\pm$ 1.77} & 83.58 \footnotesize{$\pm$ 0.84} \\
\bottomrule
\end{tabular}
}
\caption{Results of ablation study for effects of sample size and instruction strategy.}
\label{tab:ablation_summary}
\end{table}

\subsection{Human Evaluation}
\label{sec:Human_eval}
We conduct human evaluation to assess potential errors in our methodology. Specifically, we randomly selected 200 samples and recruited 3 annotators for sentence sentiment label annotation and WIS reliability scores evaluation. More details about human evaluation in Appendix \ref{sec:details_human_eval}

\subsection{Transferability to Document-level SA}
In this section, we explore whether the benefits gained from fine-tuning models on RLF sentences are transferable to document-level SA.
Table \ref{tab:document-gain} presents the document-level performance with fine-tuned LMs on the subset dataset (Section \ref{sec:Implementation}).
The results show that fine-tuned models consistently outperform zero-shot models on document-level SA both in accuracy and F1 scores. 
The non-overlapping confidence intervals indicate that these improvements are statistically significant. 
This experiment and Table \ref{tab:len-style-exp-metrics} verify that our \textbf{Lengthening} dataset enables LMs to understand better RLF with generalization ability and improvements at the word, sentence, and document levels.

\begin{table}[t]
\centering
\resizebox{\columnwidth}{!}{
\begin{tabular}{lccc}
\toprule
Backbone Model & $S_{exp}$ &  Acc(\%) & F1(\%)  \\
\midrule
\textit{\textbf{Fine-tuned with subset \& Test on subset}} \\
RoBERTa (Large) & 0.19 \footnotesize{$\pm$ .005} & 84.70 \footnotesize{$\pm$ 0.83} & 89.18 \footnotesize{$\pm$ 0.59} \\
GPT-2 (Medium) & 0.38 \footnotesize{$\pm$ .016} & 79.00 \footnotesize{$\pm$ 0.76} & 84.32 \footnotesize{$\pm$ 0.59} \\
T5 (Base) & 0.12 \footnotesize{$\pm$ .003} & 79.52 \footnotesize{$\pm$ 0.76} & 84.97 \footnotesize{$\pm$ 0.65} \\
\midrule

\textit{\textbf{Zero-shot with OOD (randomly sample 3k)}} \\
GPT-4  & 0.35 \footnotesize{$\pm$ .008} & 87.55 \footnotesize{$\pm$ 0.92} & 91.61 \footnotesize{$\pm$ 0.53} \\
LLaMA2 (13B) & 0.19 \footnotesize{$\pm$ .004} & 77.33 \footnotesize{$\pm$ 0.18} & 87.22 \footnotesize{$\pm$ 0.12} \\
\textbf{ExpInstruct} &  0.33 \footnotesize{$\pm$ .010} & 90.80 \footnotesize{$\pm$ 0.93} & 94.03 \footnotesize{$\pm$ 0.60} \\
\bottomrule
\end{tabular}
}
\caption{Comparision performances on the subset, PLMs finetuned on Lengthening and test on subset}
\label{tab:gen_res}
\end{table}

\subsection{Ablation Study}
\noindent \textbf{Effect of Sample Size} 
We compare the performance of ExpInsturct with different training sample sizes. As shown in Table \ref{tab:ablation_summary}, the best result is achieved with 1600 samples for instruction tuning, which is the data split strategy we chose for the main experiment.
We observe a continuous increase in all metrics as the sample size increases, yet the rate of gain slows after the sample size reaches 1000.

\noindent \textbf{Effect of Instruction Strategy} 
We explore how our proposed ExpInsturct strategy helps LLMs improve both performance and explainability and present results in Table \ref{tab:ablation_summary}. 
Instruction with SA (ExpInstruct w/o WIS) enhances LLMs performance for SA with RLF sentences, while Instruction with WIS improves the understanding of RLF. 
Surprisingly, ExpInstruct gains extra benefits in both explainability and performance by combining these two strategies into one task.

\noindent \textbf{Generalizability of Results} 
We conduct two experiments to support the generalizability of our findings and show results in Table \ref{tab:gen_res}. 1) Fine-tune and evaluate PLMs on the 3k subset; 2) Evaluate zero-shot performance with OOD (randomly sampled 3k instances) for LLMs. 

\section{Discussion and Conclusion}
This work sheds light on an overlooked informal style - RLF by exploring answers to two research questions.
Due to the lack of existing dataset focus on RLF, we curate the \textbf{Lengthening} dataset featuring RLF grounding from 4 public datasets.
We introduce \textbf{ExpInstruct} to improve the performance and explanation of LLMs for RLF.
We further propose a unified approach to quantify the explainability of LMs for RLF. 

Our findings uncover the expressive value of RLF from document, sentence and word levels and highlight its potential for social media content analysis, where short and informal expressions prevail.
While fine-tuned PLMs achieve superior performance than zero-shot GPT-4, their understanding of RLF still needs further improvement.
In addition, our results show the advantages of \textbf{ExpInstruct}, which can improve the performance and explainability of LLMs with limited samples. 


\section*{Limitation}
We acknowledge the limitations that present opportunities for future research. Firstly, human correction can improve the quality of the samples for explainable instruction. This can further improve the performance and interpretation of instructed LLaMA2. In addition, we can instruct-tune T5 and compare it with existing results. We leave this for future work. 

While this study focuses on RLF in English, it is important to acknowledge that this informal style is also prevalent in other languages. For instance, in Spanish, words like "graciaaas" or "holaaa!!!!" are used to convey friendliness or emphasis. In Romanian, words like "daaaa" (yes) are repeated to show strong agreement, and "minunat!!!" (wonderful!!!) to express amazement or excitement. RLFs are also commonly used in daily communications in other languages such as Chinese and Arabic. Although this paper uses datasets in English, our methodologies, including dataset generation, the ExpInstruct framework, and the unified approach for explainability evaluation can be easily transferred to other languages.

\section*{Acknowledgements}
This work was supported by the National Science Foundation awards IIS-1546480 and ITE-2137846. We also thank Andrew Schneider for his valuable contributions to our project.

{\sloppy\raggedright\bibliography{ref_list}}
\newpage
\appendix
\section{Dataset Details}
\label{sec:dataset_details}

\subsection{Algorithm for Identifying RLF}

\noindent \textbf{Sentence Segmentation and Merging.}
Utilizing open-source regular expressions (regex)\footnote{\href{https://stackoverflow.com/questions/4576077/how-can-i-split-a-text-into-sentences}{Stack Overflow discussion}}, we segment documents into individual sentences. Subsequently, to prevent the formation of trivial fragments, we merge sentences comprising fewer than three words. This approach ensures we retain meaningful linguistic structures.

\begin{table}[ht]
\centering
\resizebox{\columnwidth}{!}{
\begin{tabular}{lllcccc}
\toprule
Model  & \#Dataset (k) & \#train (k)  & \#val (k)  & \#test (k) & data split & experiment \\
\midrule
PLMs & 850 & 595.2 & 84.8 &  170  & 7 : 1 : 2 & 3 runs\\
ExpInstruct & 3 & 1.6 & 0.4  & 1.0 & custom & 3-fold\\
\bottomrule
\end{tabular}
}
\caption{Details of dataset split.}
\label{tab:dataset_split}
\end{table}

\begin{table*}[ht]
\centering
\resizebox{0.7\textwidth}{!}{
\begin{tabular}{@{}lllp{8cm}@{}}
\toprule
Lengthening Style       & POS Tag & Ratio (\%) & Examples of normalized forms                                            \\ \midrule
Letter Repetitive       & ADV     & 14.41      & 'so+', 'wa+y', 'way+', 'reall+y', 'wa+y+'                                \\
Letter Repetitive       & INTJ    & 6.67       & 'hmm+', 'ah+', 'oh+', 'um+', 'aw+'                                       \\
Letter Repetitive       & others  & 3.20       & 'all+', 'to+', 'go+d', 'bu+t', 'al+'                                     \\
Letter Repetitive       & VERB    & 3.19       & 'lo+ve', 'love+', 'lo+ved', 'recomm+end', 'lo+ves'                       \\
Letter Repetitive       & ADJ     & 3.06       & 'lo+ng', 'hu+ge', 'slo+w', 'long+', 'litt+le'                            \\
Letter Repetitive       & NOUN    & 1.71       & 'boo+k', 'go+d', 'a+s', 'wa+y', 'boo+'                                   \\
Letter Repetitive       & PRON    & 0.26       & 'you+', 'me+', 'who+', 'it+', 'sh+'                                      \\
Punctuation Repetitive  & NOUN    & 29.24      & 'book!+', 'read!+', 'series!+', 'place!+', 'product!+'                   \\
Punctuation Repetitive  & ADJ     & 10.38      & 'amazing!+', 'awesome!+', 'great!+', 'good!+'             \\
Punctuation Repetitive  & ADV     & 9.07       & 'again!+', 'ever!+', 'back!+', 'here!+', 'down!+'                        \\
Punctuation Repetitive  & PRON    & 6.87       & 'it!+', 'you!+', 'it...+', 'this!+', 'them!+'                            \\
Punctuation Repetitive  & VERB    & 6.59       & 'read!+', 'work!+', 'sucks!+', 'had!+', 'go!+'                           \\
Punctuation Repetitive  & others  & 4.03       & 'for!+', 'amazon!+', 'etc...+', 'it!+'                                   \\
Punctuation Repetitive  & INTJ    & 1.31       & 'ah+', 'please!+', 'um+', 'yes!+', 'aw+'                                 \\ \bottomrule
\end{tabular}
}
\caption{Examples of normalized forms by lengthening style and POS tag.}
\label{table:normalized-forms}
\end{table*}

\begin{figure}[ht]
\centering
\includegraphics[width=\columnwidth]{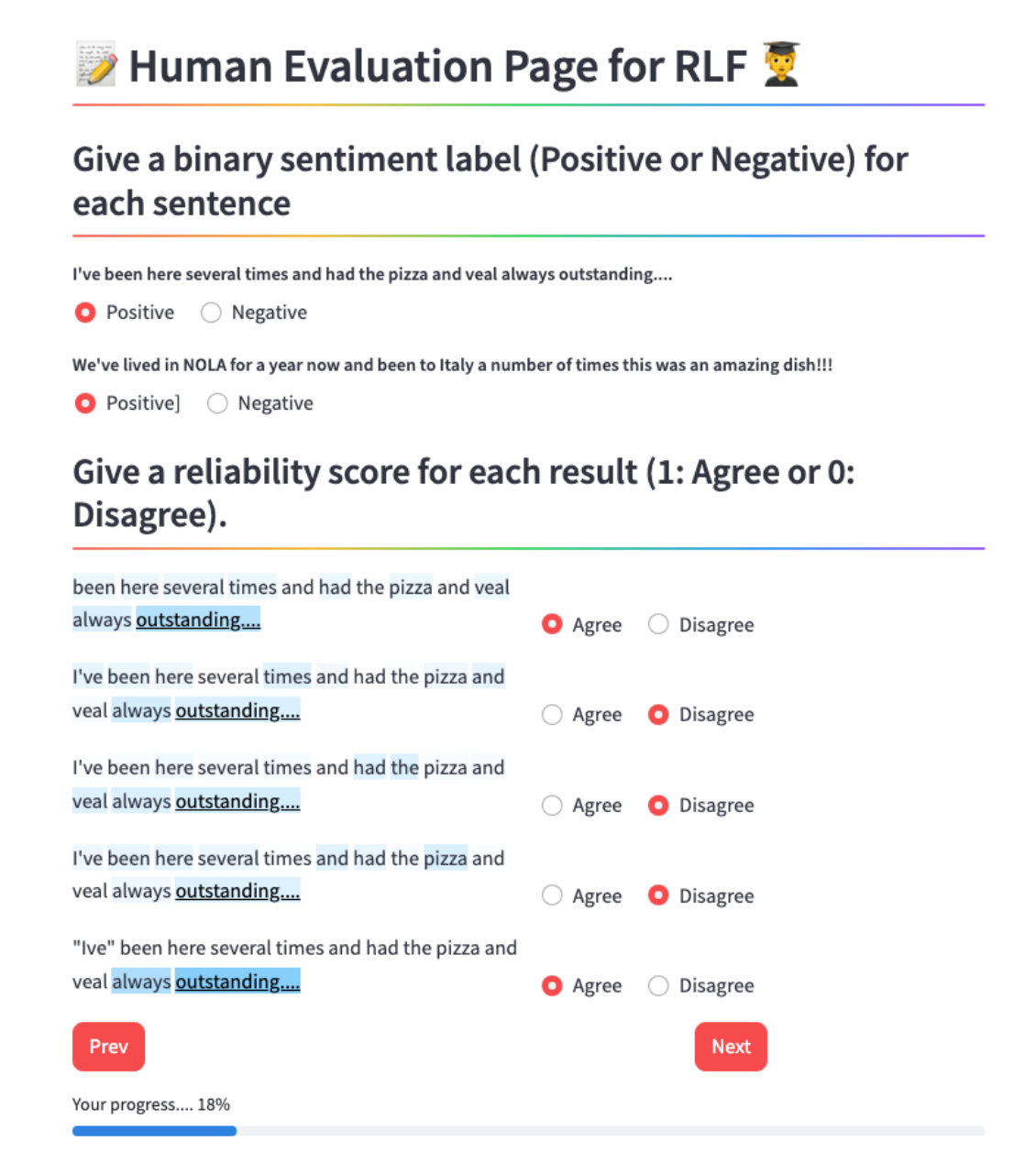}
\caption{Our customized user interface for human evaluation. Annotators are asked to do two tasks: annotation for sentiment labels and explanation reliability.}
\label{fig:human_eval_page}
\end{figure}

\noindent \textbf{Find Root Word for RLF.}
 The process involves determining the root words of these lengthened forms using a reduced method in \cite{KALMAN2014187} grounded in American English\footnote{\href{https://pyenchant.github.io/pyenchant/tutorial.html}{PyEnchant tutorial}}. For instance, variations like 'loooove' and 'loooovvve' have the generalized forms of 'lo+ve' and 'lo+v+e', and both come from the root word 'love'. We further refine our dataset by retaining only those instances where the frequency of the generalized form exceeds 100 occurrences, thereby filtering out less common variations.

\noindent \textbf{POS Tagging for RLF Words.}
Initially, we substitute each lengthened word in the sentence with its corresponding root word. Following this, we employ TweebankNLP \cite{jiang2022tweetnlp} to ascertain the part-of-speech (POS) tag of the root word. This POS tag\cite{dragut2014role} is then attributed to the respective lengthened word. We present examples of normalized forms by lengthening style and POS tag in Table \ref{table:normalized-forms}.

\noindent \textbf{Pairing RLF Sentence with w/o RLF Sentence}
For every sentence identified with RLF words, we select a corresponding w/o RLF sentence from the same document (provided it contains two or more sentences) to serve as a control sample. This allows for a comparative analysis of them. We attach the document's overall sentiment label to individual sentences extracted from it.

\noindent \textbf{Dataset Balancing}
We balance the data distribution by applying downsampling to dominant domains and lengthening styles.
Specifically, we strategically sample 20\% of sentences with repetitive letters, 8\% of those with ellipses, and all other repetitive punctuation. Additionally, to avoid data imbalance due to specific domains or generalized forms, we downsample the most prevalent ones, ensuring a more uniform distribution across the dataset.

\subsection{Dataset Split}
\label{sec:dataset_split}
The Lengthening dataset  consists of 850k samples as shown in Table \ref{tab:dataset_overview}. For our experiments with LLMs, we randomly sampled a subset of 3000 instances and ran experiments with 3-fold cross-validation strategy. In each fold, 1.6k instances were used for training, with the remaining data for validation and testing. The details of the data split are as shown in Table \ref{tab:dataset_split}.

\section{Implementation Details}
\subsection{Details of Model Parameters}
\label{sec:model_par}
All models are trained with a learning rate of 2e-5, a weight decay of 0.01, a maximum gradient norm of 1.0, and utilized gradient accumulation over 4 steps. 
We instruct-tune LLaMA2-13B-chat-hf\footnote{\href{https://huggingface.co/meta-llama/Llama-2-13b-chat-hf}{Hugging Face model card}}  with the subset samples for 5 epochs using a batch size of 1. It was set to a learning rate of 2e-4, with a reduced weight decay of 0.001 and a maximum gradient norm of 0.3, implementing a single step for gradient accumulation. Distinctively, 4-bit quantization was enabled for LLaMA2, and the compute data type was set to Float16. Moreover, the LoRA framework was integrated, set with a rank size of 64, an alpha value of 16, and a dropout probability of 0.1.
For LLaMA2, we set the temperature at 0.2, the repetition penalty at 1.4, and the maximum length to the length of the prompt plus 10 tokens. This configuration was designed to limit the size of the output token and prevent excessively long responses, thereby conserving response time and enhancing computational efficiency.

For the GPT-4\footnote{\href{https://platform.openai.com/playground?mode=chat&model=gpt-4}{OpenAI Playground}} API, we configured the temperature setting at 0.2 and established a maximum token size limit of 5,000, with other parameters remaining at default values. 

We fine-tune three PLMs. The backbones are RoBERTa (Large)\footnote{\href{https://huggingface.co/siebert/sentiment-roberta-large-english}{Hugging Face model card}}, also referred to as SIEBERT \cite{hartmann2023}, GPT-2 (Medium)\footnote{\href{https://huggingface.co/michelecafagna26/gpt2-medium-finetuned-sst2-sentiment}{Hugging Face model card}}, and T5 (Base)\footnote{\href{https://huggingface.co/mrm8488/t5-base-finetuned-imdb-sentiment}{Hugging Face model card}}.

\subsection{Analysis of RLF Styles}
In this section, we evaluate the performance of various models across two distinct styles of RLF, with results detailed in Table \ref{tab:len-style-exp-metrics}. We observe the Punctuation Repetitive style consistently achieve higher scores across three evaluation metrics and with both zero-shot and fine-tuned models, suggesting that this style is stronger for sentiment expression.

\section{Details of Human Evaluation}
\label{sec:details_human_eval}
We hire 3 graduate students as annotators for human evaluation. All annotators have fluent English levels. 
Specifically, we sample 200 instances from the subset dataset and ask annotators to conduct two annotation tasks.
We guarantee annotators receive fair wages of 20\$ per hour.

\textbf{Annotation for Sentiment Label}
Give a sentence (RLF or w/o RLF). Annotators need to give a binary sentiment label (1: Positive, 0: Negative).

\textbf{Annotation for Explanation Reliability}
We disorder and list WIS results for an RLF sentence from the 5 LMs (3 fine-tuned PLMs, ExpInstruct, and zero-shot GPT-4). Annotators need to give the reliability score for each result (1: Agree, 0: Disagree).

We customized our annotation page with streamlit\footnote{\href{https://streamlit.io/\#install}{Streamlit installation guide}} and present a case in Figure \ref{fig:human_eval_page}. Source code and sample data for this page can be found with our project link.

\end{document}